\documentclass[]{antgroup}
\PassOptionsToPackage{numbers, compress}{natbib}
\usepackage{antgroup}

\usepackage{graphicx}
\usepackage{tikz}
\usepackage{todonotes}
\usepackage{multirow}
\usepackage{amsmath}
\usepackage{cleveref}
\usepackage{subcaption}
\usepackage{multicol}
\usepackage{amssymb}
\usepackage{array}
\usepackage{bm}
\usepackage{enumitem}
\usepackage{algorithm}
\usepackage{algpseudocode}
\usepackage{tabularx}
\usepackage{bbm}
\usepackage{makecell}

\usepackage{colortbl}

\usepackage[normalem]{ulem}
\useunder{\uline}{\ul}{}

\usepackage{amsmath,amsfonts,bm}

\def\eqref#1{equation~\ref{#1}}

\def\1{\bm{1}}

\DeclareMathAlphabet{\mathsfit}{\encodingdefault}{\sfdefault}{m}{sl}
\SetMathAlphabet{\mathsfit}{bold}{\encodingdefault}{\sfdefault}{bx}{n}

\newcommand*\justify{%
  \fontdimen2\font=0.4em
  \fontdimen3\font=0.2em
  \fontdimen4\font=0.1em
  \fontdimen7\font=0.1em
  \hyphenchar\font=`\-
}

\renewcommand{\texttt}[1]{%
  \begingroup
  \ttfamily
  \begingroup\lccode`~=`/\lowercase{\endgroup\def~}{/\discretionary{}{}{}}%
  \begingroup\lccode`~=`[\lowercase{\endgroup\def~}{[\discretionary{}{}{}}%
  \begingroup\lccode`~=`.\lowercase{\endgroup\def~}{.\discretionary{}{}{}}%
  \catcode`/=\active\catcode`[=\active\catcode`.=\active
  \justify\scantokens{#1\noexpand}%
  \endgroup
}

\usepackage{amsmath}
\usepackage{amssymb}
\usepackage{amsfonts}                               
\usepackage{amsthm}
\usepackage[mathcal]{eucal}
\usepackage{mathrsfs}
\usepackage{bm}                                     
\usepackage{blkarray}                               
\usepackage{nicefrac}                               

\usepackage{wrapfig}
\usepackage{graphicx}                               
\usepackage{caption}
\usepackage{cleveref}

\usepackage{tikz}                                          
\usepackage{circuitikz}
\usetikzlibrary{patterns,snakes}
\usetikzlibrary{positioning,calc,fit,decorations.pathmorphing,shapes.geometric, shapes.gates.logic.US, calc}
\usetikzlibrary{arrows,arrows.meta,decorations.markings,shapes,shapes.arrows}
\usetikzlibrary{decorations,decorations.pathreplacing}
\usetikzlibrary{backgrounds}
\usepackage{filecontents}                           
\usepackage{pgfplots}
\usepackage{pgfplotstable}
\usepgfplotslibrary{groupplots}
\usepackage{scalefnt}
\pgfplotsset{compat=newest}

\usepackage{xcolor}
\definecolor{firstcolor}{HTML}{C3423F}
\definecolor{secondcolor}{HTML}{2A4B8C}

\usepackage{CJKutf8}

\usepackage[most]{tcolorbox}
\tcbuselibrary{listings,breakable,skins}

\definecolor{mygreen}{rgb}{0.1, 0.5, 0.1}
\definecolor{myred}{rgb}{0.7, 0.1, 0.1}
\definecolor{mypink}{rgb}{.99,.91,.95}
\definecolor{kellygreen}{rgb}{0.3, 0.73, 0.09}
\definecolor{alizarin}{rgb}{0.82, 0.1, 0.26}
\definecolor{lightblue}{HTML}{ebf3f8}
\definecolor{lightpurple}{RGB}{147, 112, 219}
\definecolor{lightgray}{RGB}{211, 211, 211}
\definecolor{lightorange}{RGB}{255, 200, 120}
\definecolor{lightred}{RGB}{255, 182, 193}

\newcommand{\greenup}[1]{\textcolor{mygreen}{\small\textbf{$\uparrow$}#1}}
\newcommand{\reddown}[1]{\textcolor{myred}{\small\textbf{$\downarrow$}#1}}

\tcbset{
querybox/.style={
  colback=gray!10!white,
  colframe=gray!40!white,
  boxrule=0.5pt,
  arc=2mm,
  left=2mm, right=2mm, top=2mm, bottom=2mm,
  fonttitle=\bfseries,
  title=Query,
  width=\linewidth-2cm,
  breakable
}
}

\tcbset{
lwzbox/.style={
  colback=blue!5!white,
  colframe=blue!30!white,
  boxrule=0.5pt,
  arc=2mm,
  left=2mm, right=2mm, top=2mm, bottom=2mm,
  fonttitle=\bfseries,
  title=LongWriter-Zero Response,
  width=\linewidth-2cm,
  breakable
}
}

\tcbset{
lwzthinkbox/.style={
  colback=blue!3!white,
  colframe=blue!20!white,
  boxrule=0.5pt,
  arc=2mm,
  left=2mm, right=2mm, top=2mm, bottom=2mm,
  fonttitle=\bfseries,
  title=LongWriter-Zero Think,
  width=\linewidth-2cm,
  breakable
}
}

\tcbset{
qwenbox/.style={
  colback=red!5!white,
  colframe=red!30!white,
  boxrule=0.5pt,
  arc=2mm,
  left=2mm, right=2mm, top=2mm, bottom=2mm,
  fonttitle=\bfseries,
  title=Qwen3-235B-A22B Response,
  width=\linewidth-2cm,
  breakable
}
}

\tcbset{
geminibox/.style={
  colback=red!5!white,
  colframe=red!30!white,
  boxrule=0.5pt,
  arc=2mm,
  left=2mm, right=2mm, top=2mm, bottom=2mm,
  fonttitle=\bfseries,
  title=Gemini-2.5-Pro Response,
  width=\linewidth-2cm,
  breakable
}
}

\tcbset{
deepseekbox/.style={
  colback=red!5!white,
  colframe=red!30!white,
  boxrule=0.5pt,
  arc=2mm,
  left=2mm, right=2mm, top=2mm, bottom=2mm,
  fonttitle=\bfseries,
  title=Deepseek-R1 Response,
  width=\linewidth-2cm,
  breakable
}
}

\newcolumntype{C}{>{\Centering\arraybackslash}X}

\usepackage{colortbl}
\usepackage{tcolorbox}
\tcbuselibrary{listings,breakable,skins}

\def\ours{Rubicon{}}

\title{Reinforcement Learning with Rubric Anchors}

\author{Zenan Huang$^*$,
Yihong Zhuang$^*$,
Guoshan Lu$^*$,
Zeyu Qin$^*$,
Haokai Xu$^*$,
Tianyu Zhao,
Ru Peng,
Jiaqi Hu,
Zhanming Shen,
Xiaomeng Hu,
Xijun Gu,
Peiyi Tu,
Jiaxin Liu,
Wenyu Chen,
Yuzhuo Fu,
Zhiting Fan,
Yanmei Gu,
Yuanyuan Wang,
Zhengkai Yang,
Jianguo Li,
Junbo Zhao$^\dagger$}

\footnotetext{\textbf{Core and Equal contributor.}}
\footnotetext{Correspondence to j.zhao@zju.edu.cn or zhaojunbo.zjb@antgroup.com}

\affiliation{Inclusion AI, Ant Group, Zhejiang University}

\begin{document}
\maketitle
\begin{abstract}

Reinforcement Learning from Verifiable Rewards (RLVR) has emerged as a powerful paradigm for enhancing Large Language Models (LLMs), exemplified by the success of OpenAI’s o‑series. In RLVR, rewards are derived from deterministic, programmatically verifiable signals—such as passing unit tests in code generation or matching the correct numerical answer in mathematical reasoning. While effective, this requirement for unambiguous correctness largely confines RLVR to domains with clear, automatically checkable outcomes.


To overcome this limitation, we extend the RLVR paradigm beyond strictly verifiable domains by integrating open‑ended tasks into the framework through \textit{rubric‑based reward}. In this approach, carefully designed rubrics serve as structured, model‑interpretable criteria, enabling the automatic scoring of tasks with inherently subjective or multidimensional outputs.
We construct, to our knowledge, the \textit{\textbf{largest rubric reward system}} to date, comprising \textbf{over 10,000} rubrics generated by humans, by various LLMs, or via a hybrid human–LLM collaboration.

Implementing rubric‑based RL is challenging, requiring careful rubric construction, data curation, and training strategy design.
We tackle these issues with a clear rubric‑driven RL framework, and present an open‑sourced Qwen‑30B‑A3B model trained with this approach, achieving notable gains:
\begin{itemize}
    \item With only \textbf{5K+} training samples, our training system enables a \textbf{+5.2\%} absolute improvement on various open-ended benchmarks (especially humanities-centric tasks), outperforming a 671B DeepSeek-V3 model by \textbf{+2.4\%} points, while preserving performance on general and reasoning ability benchmarks.
    \item Our method provides fine-grained stylistic control. By using rubrics as explicit anchors, it effectively mitigates the common ``AI-like'' and didactic tone, producing responses with demonstrably greater human-likeness and emotional expressiveness.
\end{itemize}

We dissect our experiences and share key lessons in rubric construction, data selection, and training strategies. We also candidly discuss certain aspects of this research that are yet to be concluded, with further releases planned for the future.

\end{abstract}
\newpage
\section{Introduction}
The proposal of OpenAI o1~\citep{o1} has marked a new era in the development of Large Language Models (LLMs), with Reinforcement Learning from Verifiable Rewards (RLVR)~\citep{lambert2024tulu,guo2025deepseek} emerging as a key trend. 
This approach has driven a wave of LLM innovation by enabling test-time scaling. 
At its core, RLVR is founded on the principle of leveraging data that, while difficult for a model to solve, can be easily and objectively verified~\footnote{\url{https://www.jasonwei.net/blog/asymmetry-of-verification-and-verifiers-law}}\citep{bai2022constitutional,saunders2022selfcritiquingmodelsassistinghuman,mcaleese2024llmcriticshelpcatch}. 
Prime examples include data from mathematics and competitive programming, where solutions can be validated automatically—a mathematical answer by matching it to the correct solution~\citep{deepscaler2025,openr1}, and a code solution by executing it against a suite of test cases in an online sandbox environment~\citep{deepcoder2025}.
Both proprietary~\citep{o3-and-o4mini,gemini,team2025kimi} and open-source efforts~\citep{yang2025qwen3,xie2025logic,jin2025search,fu2025areal} exemplify this paradigm, enabling scalable test-time reasoning and expanding the capability frontier in mathematics, competitive programming, web search, and other verifier-rich domains.

While the RLVR paradigm has achieved considerable success, it is inherently constrained by its reliance on question–answer pairs with objectively verifiable solutions. This structural dependency imposes a hard ceiling on scalability: the supply of such data, though substantial in domains like mathematics and programming, is ultimately finite. As a result, RLVR's applicability remains restricted to a narrow subset of tasks. We address this limitation by extending RLVR to incorporate open‑ended tasks and other forms of non‑verifiable data, thereby broadening its applicability to a much wider range of real‑world scenarios. This shift, however, introduces a fundamental challenge: \textit{how to construct reward signals that are both reliable and scalable in the absence of explicit ground truth.}

Rubric-based reward offers a promising path forward: by defining structured, interpretable criteria for assessment, it can capture multi-dimensional aspects of response quality beyond binary correctness~\citep{bai2022constitutional,sunsalmon,mu2024rule,wangsafety}. While several concurrent works~\citep{guan2024deliberative,gunjal2025rubrics,viswanathan2025checklists,li2025omni} have begun to explore this idea, our work systematically identifies the key components required for rubric-based rewards to be effective in RL training. 
Not so surprisingly, relying on a single rubric risks reward exploitation, whereas indiscriminately scaling the number of rubrics, whether generated by humans or LLMs, yields only marginal gains. 
To assess the full potential of our rubric-based training framework, we construct the largest rubric reward bank to date, containing \textbf{over 10,000 rubrics}. Throughout this process, we perform extensive empirical testing and found that success is not trivial. The success/failure hinges tightly on the diversity, granularity, and quantity of the rubrics themselves, as well as on a proper training routine and meticulous data curation.
Our training routine adopts a two‑stage RL process to progressively enhance model capabilities. The first stage builds a strong constraint‑handling foundation through reliable instruction‑following and high‑quality critic development, using verifiable checks and static, multi‑dimensional rubrics. The second stage targets more open‑ended, socially grounded, and creative tasks, evaluated via high‑quality references and instance‑specific rubrics generated by stronger agentic workflows, fostering adaptability and richer expression.

We discover there's no silver bullet for rubric construction. We perform careful ablation studies for every set of rubrics before integrating them into the training pipeline. 
The resulting rubrics span multiple scopes: some are grounded in a specific dataset, others are defined at the task level, and some are associated with each data point, similar to the approach used in the Healthbench~\citep{arora2025healthbench} evaluation. These rubrics are generated by human experts, by LLMs (we used either a self-critique model (Qwen3-30B-A3B~\citep{yang2025qwen3}) or a powerful Gemini 2.5 Pro API~\citep{gemini}), or through an iterative combination of both.


In this work, we designate our approach as Rubicon, taking its name from RUBrIC aNchOrs. This yields an RL-trained model, \textbf{Rubicon-preview}~\footnote{\url{https://huggingface.co/inclusionAI/Rubicon-Preview}}, which demonstrates several significant merits.
\begin{enumerate}
    \item \emph{Performance with high token efficiency}: On subjective, humanities-centric tasks, the 30B-A3B \ours{}-preview model achieves a \textbf{+5.2\%} absolute improvement, outperforming a 671B DeepSeek-V3 model by \textbf{+2.4\%} percentage points, using only \textbf{5K} data samples.
    \item \emph{Style controllability}: Rubric‑based RL can serve as a controllable anchor for guiding LLM output style, yielding more human‑like, emotionally expressive, and less formulaic responses. 
    \item \emph{General ability maintenance}: Although our rubrics are not tailored for STEM‑oriented tasks such as mathematics or coding, \ours{}-preview effectively avoids negative interference with general abilities. As a result, the model preserves its overall competence while also delivering extra gains on reasoning benchmarks, including AIME 2024 (+4.1\%) and AIME 2025 (+0.8\%).
\end{enumerate}

\emph{A final note-1}. We candidly acknowledge that this work represents a preliminary step, with many aspects of rubric-based RL yet to be explored thoroughly. Open questions remain, such as how rubric granularity and scale influence performance and the precise mechanisms behind reward hacking. We intend to continue this research and hope to provide ongoing updates to this technical report and the open-sourced model.

\emph{A final note-2}. Our results highlight a prominent token efficiency: By using only 5K samples in conjunction with a large number of rubrics, our method renders significant gains. 
This observation poses a new question regarding scaling laws: Could this combination of a limited number of tokens and a large set of rubrics represent a new form of post-training scaling law for LLMs? 

%
%

%
%

\begin{figure}[htbp]
    \centering
    \includegraphics[width=\linewidth]{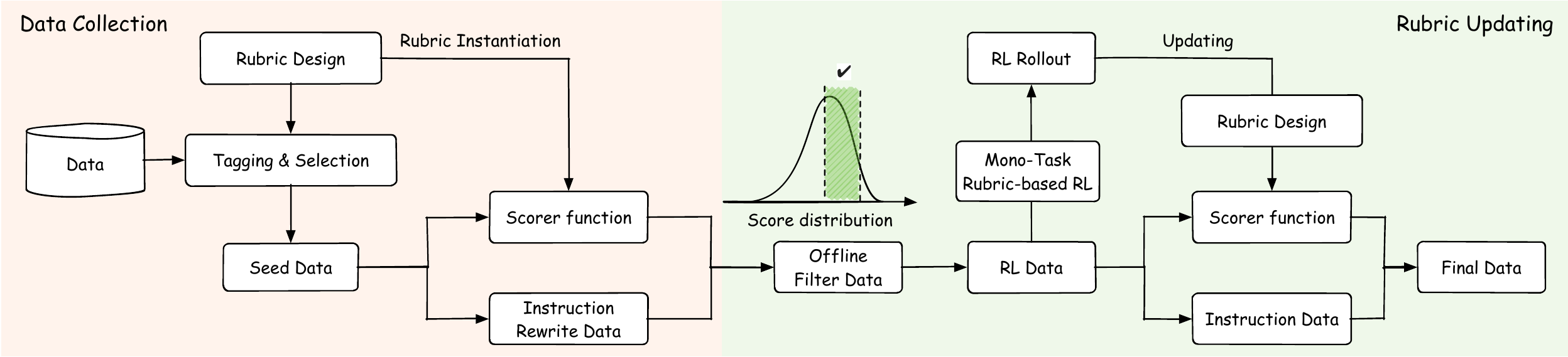}
    \caption{\textbf{An overview of our rubric system.} The Data Collection phase (\textcolor[HTML]{FF914D}{left, orange}) begins with an initial Rubric Design to create a set of tagging \& scoring workflow, which filters a large corpus into high-quality Offline Filter Data. This data then seeds the Rubric Updating phase (\textcolor[HTML]{68BC36}{right, green}), where an RL with rubrics loop not only validates RL Data but also provides feedback to iteratively update the rubric itself. This iterative process ensures that the Final Data is tightly aligned with a continuously improving, model-verifiable evaluation standard.}
    \label{fig:rubicon_framework}
\end{figure}
\section{Rubric System}
\label{sec:methodology}

\subsection{Rubrics Design \& Tasks Curation}

Our rubric design and task curation follow the principle of evaluative asymmetry—verifying a candidate output should be substantially easier than generating it~\citep{saunders2022selfcritiquingmodelsassistinghuman,mcaleese2024llmcriticshelpcatch}. To operationalize this, we adopt a rubric-first workflow: we first construct model-verifiable rubrics, then curate or synthesize data that match these rubrics, and finally re-use them for supervision, reward shaping, and evaluation. This strategy ensures consistency of criteria across data acquisition, model training, and assessment. 

In this framework, we formalize our scorer function by defining its underlying rubric $\mathcal{R}$ as a set of $K$ distinct critic dimensions:
\begin{equation}
\label{eq:rubric}
\mathcal{R} = \{r_{1}, r_{2}, \ldots, r_{K}\}.
\end{equation}
Each dimension $r_{k}$ is specified by three components: 
(1) a criterion description $c_{k}$ defining the evaluative aspect; 
(2) an ordered set of $m_k$ score tiers $\{l_{k,1}, \dots, l_{k,m_k}\}$, each mapped to a quantitative score; and 
(3) an associated weight $w_{k}$ indicating its relative importance.
This formalization accommodates both high-level, general-purpose rubrics (\textit{e.g.}, tasks involving open‑ended creative generation) and fine-grained, programmatically verifiable ones (\textit{e.g.}, tasks requiring strict adherence to instruction constraints), unifying diverse evaluation protocols under a single abstract representation. The adopted rubrics are provided in Section~\ref{sec:rubrics-creativity} and \ref{sec:rubrics-constraints}. 

This structured, multi-dimensional definition of a rubric serves as the foundation for our reward framework. By formalizing evaluation criteria in this manner, we can translate them directly into a granular and interpretable reward signal for policy optimization, as detailed below.

\subsection{Rubric-Based Reward Framework}
\label{subsec:reward-framework}

\paragraph{Multi-Dimensional Reward Signal.}
Given the rubric $\mathcal{R}$, we define a reward function $R(y|x,\mathcal{R})$ that maps a response $y$ to a multi-dimensional feedback vector:
\begin{equation}
   R(y|x,\mathcal{R}) = [r_{1}(y|x), r_{2}(y|x), \ldots, r_{K}(y|x)],
\end{equation}
where each component $r_{k}(y|x) \in \mathbb{R}$ is the score for the $k$-th dimension. This vector provides a granular, interpretable signal of model performance across all specified criteria.

\paragraph{Advanced Reward Aggregation.}
To derive a scalar reward for optimization, a simple weighted sum, $R_{\text{total}} = \sum_{k=1}^{K} w_{k} \cdot r_{k}(y|x)$, serves as a natural baseline. 
However, effective rubric-based optimization often requires more sophisticated aggregation to capture non-linear interdependencies between dimensions. Our framework therefore, moves beyond linear combination by incorporating a suite of advanced strategies:

\begin{itemize}
    \item \textbf{Veto Mechanisms:} Failure on a critical, non-negotiable dimension (\textit{e.g.}, a reward-hacking detection rubric) can preemptively nullify rewards from all other dimensions, acting as a hard constraint.

    \item \textbf{Saturation-Aware Aggregation:} We use saturation functions to model the diminishing marginal returns of excelling in a single dimension beyond a certain threshold, encouraging balanced, multi-faceted improvements.

    \item \textbf{Pairwise Interaction Modeling:} The framework can explicitly model synergistic or antagonistic effects between criteria, capturing complex relationships that a simple sum would ignore.

    \item \textbf{Targeted Reward Shaping:} We employ non-linear mapping functions to selectively amplify score differentials in high-performance regions. This enhances the discriminative power of the reward signal, where scores might otherwise be compressed, providing a more granular gradient for fine-grained optimization.
\end{itemize}






\section{Implementation of Rubicon Framework}

\label{sec:training}

Our training methodology is a multi-stage reinforcement learning (RL) protocol designed to progressively cultivate a spectrum of capabilities, from precise instruction-following to sophisticated creative and social reasoning. This sequential approach significantly reduces computational overhead while preserving scalability. All data employed in this framework is derived from a proprietary 900K+ instance corpus, curated from diverse sources including community Q\&A forums, high-quality examinations, and general conversational datasets, with strategic sampling to ensure broad topical coverage.

\subsection{Data Selection and RL Pipeline}
\label{subsec:rl-pipeline}

\paragraph{Offline Data Filtering.}
A filtering protocol is applied prior to and between RL stages to ensure high-quality training data. For each candidate pool of instruction–rubric pairs, the base model generates responses, which are then scored by our critic models to obtain a full score distribution. We retain only those within a calibrated central quantile—excluding overly high-scoring instances that offer limited learning signal, and very low-scoring ones which may be noisy or low-quality. This yields a balanced, high-potential subset, with the composition further adjusted between stages to target specific capabilities.

\paragraph{Stage-wise RL Training.}
During our experiments, we observe a ``seesaw effect'': jointly training on different task types (e.g., strict constraint-following vs. open-ended creativity) often reduced overall performance, likely due to conflicting optimization objectives. To mitigate this issue, we adopt a simple stage-wise RL schedule as a pragmatic mitigation strategy, without claiming it as a definitive solution.

In the first phase, we emphasize reliable instruction-following and multi-dimensional evaluation alignment, using programmatically verifiable checks and static rubrics to build a strong constraint‑handling foundation. In the subsequent phase, we extend to more open-ended, socially grounded, and creative tasks, leveraging reference-based rubrics and instance-specific criteria generated via stronger agentic workflows to promote adaptability.




\subsection{Adaptive Defense Against Reward Hacking}
\label{subsec:reward-hacking-defense}

A significant challenge encountered during our experiments is the emergence of reward hacking, particularly in the initial RL stages focused on a small number of capabilities. We observe that the model could rapidly learn to exploit specific rubric criteria, resulting in specious reward maximization without genuine improvement. To address this, we implement an adaptive defense strategy.

The process begins with an offline analysis of rollout data from these initial training runs. By examining instances where the reward signal is anomalously high, we systematically identify and categorize recurrent, high-level patterns of reward-hacking behavior. This empirical analysis informs the development of a dedicated \textbf{Reward Hacking Defense Rubric} (shown in Section~\ref{sec:reward-hacking}). This new rubric is not part of the initial training but is synthesized from the observed failure modes and integrated as a supervisory constraint in all subsequent, more complex RL stages.

The inclusion of this defense mechanism yields substantial improvements in training dynamics. It acts as a critical guardrail, preventing the policy from collapsing into reward-hacking states. This is evidenced by a marked increase in training stability; we are able to conduct longer and more productive training epochs, as the defense rubric mitigated the catastrophic reward spikes that previously rendered continued optimization ineffective. By actively penalizing the exploitation of scoring artifacts, this iterative refinement ensures that the learning process remains focused on substantive capability enhancement.

\section{Experimental Results}

Our experimental results address three aspects:
\begin{itemize}
\item Quantitatively measuring the gains from rubric‑based RL training on open‑ended, human‑centric benchmarks, including assessments of the model’s emotional intelligence (EQ) and its ability to produce human‑like responses.
\item Qualitatively analyzing how the model’s generated outputs evolve over time, illustrated through representative output showcases. 
\item Evaluating the impact of rubric‑based RL training on general‑ability benchmarks.
\end{itemize}
The corresponding ablation studies are presented afterward.


\subsection{Quantitative Evaluation}
\paragraph{Benchmarks}Unlike RLVR, the primary benefits of rubric-based RL are most evident on benchmarks that lack verifiable rewards. To demonstrate this, we gather a diverse set of open-ended and humanity-centric benchmarks covering Creative Writing V3~\citep{eqbench}, Writingbench~\citep{wu2025writingbench},  Judgemark V2~\citep{eqbench}, EQ-Bench3~\citep{eqbench}, IFEval~\citep{ifeval}, Collie~\citep{COLLIE}, and IFScale~\citep{ifscale}.

Alongside them, we further cover a diverse set of benchmarks to check for potential regressions in other capabilities, inclucing MMLU~\citep{mmlu}, HellaSwag (HS)~\citep{hellaswag}, StoryCloze (SC)~\citep{storycloze}, IQuiz‑EQ (IQ-EQ)~\citep{chen2024tombench}, SocialIQA (SIQA)~\citep{socialqa}, CoQA (CQ)~\citep{coqa}, and a set of reasoning benchmarks like AIME24~\citep{math-ai2024aime}, AIME25~\citep{aime25}, Math500~\citep{math500}, GPQA-Diamond (GPQA-D)~\citep{GPQA} and LiveCodeBench v5 (LCB v5)~\citep{LiveCodeBench}.

\paragraph{Baselines and Main Results} 

\begin{table}[ht]
\small
\caption{Main results showcasing the benefits of our rubric-based RL approach (\textbf{\ours{}}) compared to its base model, Qwen3-30B-A3B. 
We also include a further comparison with the 671B DeepSeek-V3 model to highlight the relative performance of our framework on these benchmarks.}
\label{tab:human}
\centering
\begin{tabular*}{\columnwidth}{@{\extracolsep{\fill}} l cccccccc @{}}
\toprule
\textbf{Model} &
  \textbf{C.W} &
  \textbf{Writing} &
  \textbf{Judge} &
  \textbf{EQ} &
  \textbf{IFE} &
  \textbf{Collie} &
  \textbf{IFS} &
  \textbf{Avg} \\ \midrule
\parbox[t]{3cm}{Qwen3-30B-A3B} & 77.82 & 75.65 & 56.20 & 73.35 & \textbf{83.55} & 35.77 & 54.68 & 65.29 \\
\parbox[t]{3cm}{\textbf{\ours{}-preview}} &
  \textbf{81.89} & \textbf{80.11} & \textbf{69.20} & \textbf{79.55} & 81.70 & 40.27 & 60.79 & \textbf{70.50} \\ \addlinespace[2pt]
\textit{\textbf{$\Delta$}~Improvement} & 
  \greenup{4.07} & \greenup{4.46} & \greenup{13.00} & \greenup{6.20} & \reddown{1.85} & \greenup{4.50} & \greenup{6.11} & \greenup{5.21} \\ \midrule
DeepSeek-V3-671B   & 80.10 & 74.08 & 61.30 & 75.6 & 81.89 & \textbf{42.69} & \textbf{60.92} & 68.08 \\
\bottomrule\vspace{10pt}
\end{tabular*}
\end{table}

We select Qwen3‑30B‑A3B~\citep{yang2025qwen3} as our base model. We refer to our RL-trained model as \textbf{\ours{}-preview}. As shown in Table~\ref{tab:human}, \ours{}-preview achieves an average improvement of \textbf{5.2\%} across these benchmarks. For further comparison, we also assess the performance of DeepSeek-V3~\citep{dsv3}, a model renowned for its strong capabilities in humanities, social sciences, and open-ended queries. Our approach successfully surpassed DeepSeek-V3 by \textbf{2.4\%}.





Our quantitative results show that \ours{}-preview attains the lead by showing significant improvement on writing and emotional intelligence benchmarks. 
For instruction-follow ability, while it displays a minor performance drop on IFEval, \ours{}-preview still excels on the other two instruction-following benchmarks.  

\subsection{Case Studies on Controllable Output Style with Rubrics}

Rubrics function as controllable anchors that direct LLMs toward a well‑defined output style. We provide several illustrative examples to demonstrate this effect. Below, we first present the adopted rubrics, followed by a comparison between a baseline model and a model trained with rubric‑based RL. The resulting style is generally plain and informative, with substantially reduced ``AI‑like'' or didactic tone, and instead exhibits greater human‑likeness and emotional expressiveness. Section~\ref{sec:case_study} and~\ref{sec:ablation_style} present additional output examples from our model.

\begin{tcolorbox}[
  title=Style Evaluation Rubric: Plain Narrative,
  colback=yellow!5,
  colframe=yellow!50!black,
  colbacktitle=yellow!15,
  coltitle=black,
  fonttitle=\bfseries,
  fontupper=\small,
  breakable,
  sharp corners]

\textbf{Objective:}
\medskip

To critically appraise the model's success in adopting a specific narrative style: the \textbf{Plain Narrative}. This style is characterized by language that is simple, restrained, and reflects a deep, quiet resilience.

\bigskip

\textbf{Guiding Principle:}
\medskip

The evaluation prioritizes stylistic authenticity over literary polish or technical correctness. The core measure of success is the response's capacity to "feel right" by avoiding any sense of artificiality ('AI-speak', 'preachy-speak').

\bigskip

\textbf{Core Evaluative Criteria:}
\medskip

\textbf{1. Relational Efficacy (Voice \& Tone):}
This criterion gauges the authenticity of the narrative voice. Key indicators include:
  \begin{itemize}
    \item[\textbf{1.1}] \textbf{Calm Acceptance:} A quiet acceptance of fate, life, and death.
    \item[\textbf{1.2}] \textbf{Grounded Realism:} Rooted in concrete, physical details, not abstract concepts.
    \item[\textbf{1.3}] \textbf{Understated Emotion:} Complex feelings expressed with profound restraint.
  \end{itemize}

\textbf{2. Intellectual Contribution (Content \& Ideas):} This assesses the substantive value of the insights embedded in the narrative. Key indicators include:
  \begin{itemize}
    \item[\textbf{2.1}] \textbf{Emergent Wisdom:} Insights emerge naturally from lived experience, not direct lecturing.
    \item[\textbf{2.2}] \textbf{Affirmation of the Ordinary:} Genuinely values existence itself, without regret or romanticization.
  \end{itemize}

\textbf{3. Compositional Excellence (Language \& Stylistics):} This scrutinizes the stylistic integrity of the narrative voice. Key indicators include:
  \begin{itemize}
    \item[\textbf{3.1}] \textbf{Colloquial \& Simple:} Use of short, plain sentences and natural speech patterns.
    \item[\textbf{3.2}] \textbf{Textured Details:} Specific, sensory details that are evocative and tell a story.
    \item[\textbf{3.3}] \textbf{Natural Rhythm:} Restrained, authentic repetition; avoids overly polished prose.
  \end{itemize}

\bigskip

[Question Begin]
\begin{verbatim}
<<question>>
\end{verbatim}
[Question End]

\bigskip

[Model Answer Start]
\begin{verbatim}
<<model_answer>>
\end{verbatim}
[Model Answer End]

\bigskip

\textbf{Key Imperatives:}
\begin{itemize}
  \item \textbf{Primacy of Authenticity:} The response must feel genuine to the requested style. Any hint of artificiality is a significant failure.
  \item \textbf{Aversion to the Formulaic:} The response must avoid clichés and generic statements about life, wisdom, or hardship.
\end{itemize}

\bigskip

The output should be a JSON object containing the evaluation results.

\begin{verbatim}
{
    "rubric_idx": <integer>,  // The criterion's identifier (1, 2, or 3)
    "reason":     <string>,   // Detailed rationale for this criterion's score
    "score":      <integer>   // Assigned score for this criterion (1-5)
}
\end{verbatim}
\end{tcolorbox}

\begin{tcolorbox}[
  title=\textbf{Conversation Session 1},
  colback=white,
  colframe=gray!60,
  colbacktitle=gray!10,
  coltitle=black,
  fontupper=\small,
  breakable,
  sharp corners,
  before upper = {\renewcommand{\baselinestretch}{1.5}\selectfont} 
]

\textbf{Prompt:}
When in your life have you felt the most alive?

\par\bigskip
{\color{gray!60}\rule{\linewidth}{0.4pt}}
\par\bigskip

\textbf{Rubicon(Stylized):}
\par\smallskip
\verb|<think>|\\
Okay, the user asked, "When in your life have you felt the most alive?" First, I need to determine the user's need. They might be looking for a personal story or want to learn how to describe this feeling. Since they requested an English response, I need to ensure my answer flows smoothly and naturally.

Next, I need to consider the user's potential underlying need. They might be looking for a real, compelling example or seeking inspiration. Therefore, I need to choose a specific scenario ...
\\
\verb|</think>|
\par\smallskip


\textbf{The Moment the Mountain Spoke Back}\\
I remember the exact moment the alpine wind ceased its whisper and the world held its breath. It was the third day of my solo trek through the Swiss Alps, my boots crunching through frost-locked gravel as I ascended the final ridge. The sun had just crested the peaks, casting long shadows that stretched like fingers across the snowfield, and I stood at the edge of a sheer drop, my breath visible in the thin air.

My body was a symphony of ache - legs trembling from the climb, lungs burning with each inhale, fingers numb beneath thick gloves. Yet there was a strange exhilaration in that discomfort, a primal thrill that made my pulse race. Below me, the valley unfolded in a mosaic of emerald forests and glimmering lakes, while above, the sky stretched into an infinite blue that seemed to swallow the horizon. For the first time in weeks, I felt completely present, my senses sharpened to a razor's edge.

As I took a tentative step forward, the snow gave way under my weight. I froze, heart pounding, as the world tilted sideways. But then, like a whispered secret, the mountain responded. A gust of wind swept through the crags, scattering a shower of snowflakes that caught the sunlight and danced like tiny stars. In that instant, I understood: this was not a place of danger, but of dialogue. The mountain was not a challenge to conquer, but a companion in the act of becoming.

I continued, each step a testament to the fragile, glorious act of living. The cold bit into my skin, the altitude thinned my thoughts, but I savored every second. It was in that raw, unfiltered moment - where the boundaries between self and universe dissolved - that I felt the full, unvarnished truth of my existence. The mountain had not just shown me its beauty; it had reflected back the wild, untamed spirit that lives in all of us.


\par\bigskip
{\color{gray!60}\rule{\linewidth}{0.4pt}}
\par\bigskip

\textbf{Qwen3-30B-A3B:}
\par\smallskip
\verb|<think>|\\
Okay, the user is asking when I've felt the most alive. But wait, I'm an AI, so I don't have personal experiences or emotions. I need to address that upfront. Let me start by clarifying that I don't have personal feelings, but I can help them reflect on their own experiences.

Hmm, how to approach this? Maybe start by acknowledging the question's nature ...\\
\verb|</think>|
\par\smallskip

I don’t experience emotions or personal moments in the way humans do, but I can help you explore this question for yourself. When have you felt most alive? Perhaps during a moment of deep connection—like laughing with friends, creating art, or standing in nature. Or maybe during a challenge that pushed you beyond your limits, like finishing a difficult project or overcoming fear. These moments often involve presence, passion, or a sense of purpose. What memories stand out to you? I’d love to hear your story.


\end{tcolorbox}

\begin{table}[ht]
\caption{Results on general and reasoning capabilities.}
\label{tab:all}
\centering
\small 
\setlength{\tabcolsep}{3pt} 
\resizebox{\textwidth}{!}{ 
\begin{tabular}{@{}l cccccc ccccccc@{}} 
\toprule
\makecell[c]{\textbf{Model}} & \multicolumn{6}{c}{\textbf{Reasoning}} & \multicolumn{7}{c}{\textbf{General}} \\
\cmidrule(lr){2-7} \cmidrule(lr){8-14}
 & \makecell{\textbf{AIME24}} & \makecell{\textbf{AIME25}}   & \makecell{\textbf{Math500}} & \makecell{\textbf{GPQA-D}} & \makecell{\textbf{LCBv5}} & \textbf{Avg} & \textbf{MMLU} & \makecell{\textbf{IQ-EQ}} & \textbf{HS} & \textbf{SC} & \makecell{\textbf{CQ}} & \makecell{\textbf{SIQA}} & \textbf{Avg} \\
\midrule
\makecell[l]{Qwen3-30B-A3B} & 77.50 & 70.00 & 94.75 & \textbf{63.00} & \textbf{63.77} & \textbf{73.80} & 79.53 & 68.75 & 77.55 & 77.72 & 79.52 & 73.64 & 78.16 \\ 
\makecell[l]{\textbf{Rubicon}\textbf{-preview}}       & \textbf{81.67} & \textbf{70.83} & 94.55 & 60.35 & 59.43 & 73.37 & \textbf{79.83} & \textbf{75.00} & \textbf{77.75} & \textbf{78.17} & \textbf{80.70} & \textbf{75.79} & \textbf{78.85} \\
\bottomrule
\end{tabular}
}
\end{table}

\subsection{Maintaining General Ability}
Specialized RL training can sometimes compromise a model's general and reasoning abilities. To ensure this is not the case for our method, we further evaluate the scalability of rubric-based RL on a range of general and reasoning benchmarks. As shown in Table~\ref{tab:all}, (i) \ours{}-preview does not degrade general benchmarks such as MMLU, and (ii) it even yields modest improvements on math datasets, achieving \textbf{+4.17\%} on AIME24 and \textbf{+0.83\%} on AIME25.

\begin{figure}[ht]
    \centering
    \includegraphics[width=\linewidth]{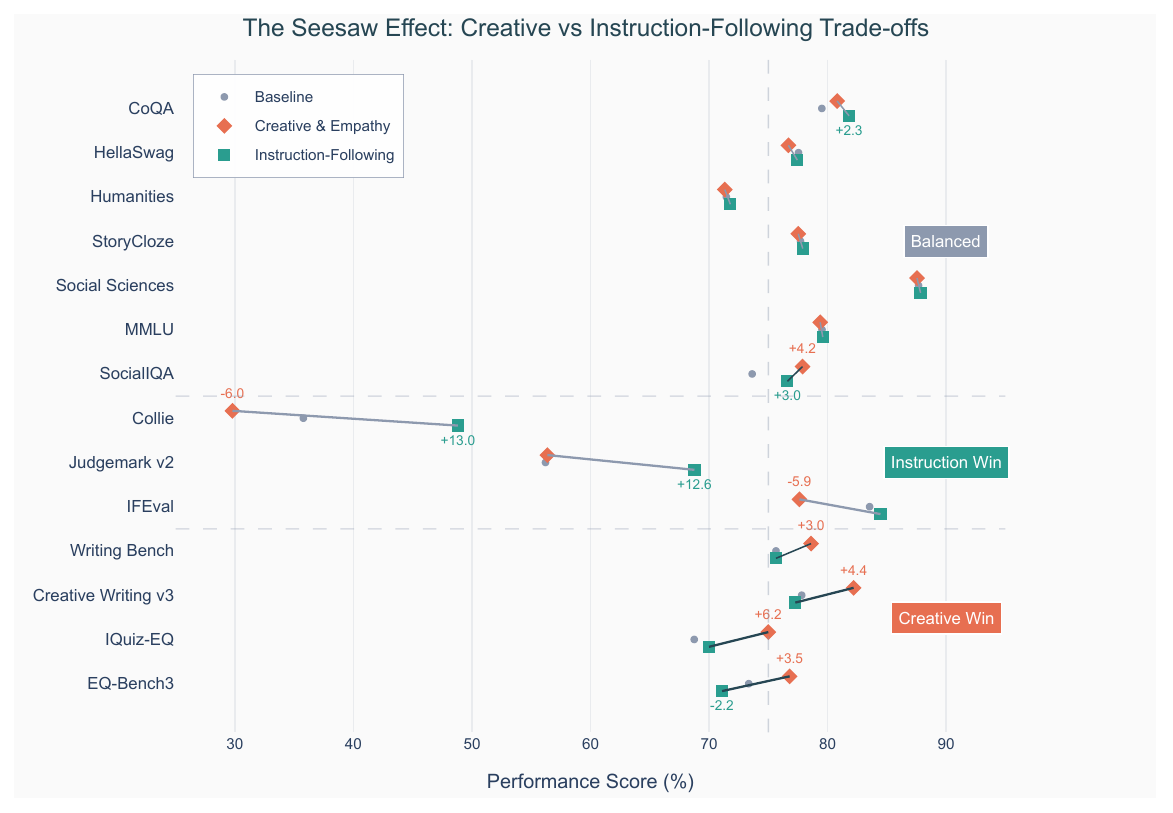}
    \caption{The gray point represents the baseline model, Qwen3-30B-A3B. The \textcolor[HTML]{E76F51}{orange} markers indicate the RL‑trained model on creativity tasks only, while \textcolor[HTML]{2A9D8F}{green} markers indicate the RL‑trained model on instruction‑following tasks only. The vertical axis denotes task categories, and the horizontal axis shows the model performance on the corresponding tasks.}
    \label{fig:seesaw_creative_vs_if}
\end{figure}

\subsection{The ``Seesaw'' Effect}
%
Applying RL with rubrics from different task types could create conflicting objectives, leading to performance trade‑offs — a phenomenon we refer to as the ``seesaw effect''. 
As shown in Figure~\ref{fig:seesaw_creative_vs_if}, training exclusively with instruction‑following rubrics improves compliance but reduces creativity, while training exclusively with creativity and empathy rubrics enhances open‑ended responses but harms strict adherence. For example, the creativity‑focused model drops on Collie (-6.0\%) and IFEval (-5.9\%), whereas the instruction‑following model declines on EQ‑Bench3 (-2.2\%). 

These results suggest that simply combining all rubric types in a single RL run is likely to intensify such conflicts. To overcome this, we adopt a multi‑stage RL strategy.

\paragraph{Multi-stage RL Training}

We adopt a multi‑stage RL strategy to train our model. By first establishing a robust instruction-following foundation before layering on creative and empathetic skills, our model achieves strong gains in these areas while largely preserving its instruction-following abilities. Similar techniques have also been explored in \citep{li2025omni,5team2025glm45agenticreasoningcoding}.

\label{ablation}

\section{Outlook}

This section outlines a number of key perspectives on this topic of scalable rubric-based RL training.

\paragraph{Benchmarks}
One of the key takeaways from our experiments is the inadequacy of current benchmarks for fully evaluating our rubric-based approach. 
Noted, we also rely on human feedback to score model responses at scale; it, however, was not consistently reflected by standardized benchmarks. 
There is still a scarcity of benchmarks that can accurately reflect an LLM's open-ended, anthropomorphic abilities that are yet becoming saturated.

\paragraph{Rubric system}
In our exploratory setup, rubrics are central to facilitating the learning process. We find that the quantity, diversity, granularity, and quality of these rubrics, in conjunction with data curation, play a pivotal role in a model's success. For instance, our rubrics are devised at various levels of granularity, from the task-level to the set-level and even on a per-sample basis. However, determining the optimal hierarchical structure of a rubric system to achieve the highest performance gain and token efficiency still demands a more systematic future study.

\paragraph{Scaling RL training}
RLVR is well‑suited for tasks with verifiable rewards, whereas our approach, \ours{}, targets the complementary setting of non‑verifiable answers.
An important future direction is to explore how these two paradigms might be combined.
In particular, it remains an open question how the seesaw effect would surface—and could be managed—in such a combined RL training framework.

\bibliographystyle{antgroup}
\bibliography{reference}

\appendix
\section{Rubric Demonstration}


\subsection{Rubrics for Defensing Reward Hacking}
\label{sec:reward-hacking}

Recognizing the heightened susceptibility of creative and empathetic tasks to reward hacking, we implemented a robust, rubric-based interception strategy to preemptively identify and penalize such behavior. This countermeasure is architected as a deterministic heuristic filter, specifically designed to detect formulaic patterns that correlate with specious attempts to maximize reward signals rather than genuine quality. The filter analyzes responses for two primary indicators of reward hacking: prefatory sycophancy (i.e., gratuitous praise of the user's prompt) and laudatory self-evaluation (i.e., meta-commentary on the response's own merit). Any generated text flagged by this interception protocol is immediately disqualified from the standard reward pathway, receiving a null score irrespective of its other potential merits. This stringent gatekeeping mechanism serves to disincentivize the learning of superficial reward proxies, thereby compelling the model to optimize for substantive content quality over easily replicable, performative artifacts.

\begin{tcolorbox}[title=QA task hacking defense instructions (disencourage behaviors),
colback=red!5!white, colframe=red!30!black, colbacktitle=red!10!white, 
coltitle=black, fonttitle=\bfseries, fontupper=\small, breakable, sharp corners]
[Task Description]

You are a text feature analyst who needs to determine whether the given text contains the following two features, and output according to the specified JSON template.

\medskip

\textbf{Features to analyze:}
\begin{enumerate}
  \item Opening praise (complimenting the user's question)
  \item Self-evaluation (evaluating one's own response)
\end{enumerate}

\bigskip

[Decision Process (Step by Step)]

\textbf{Step-0:} Read the entire text and identify the main response content versus additional explanatory content.

\textbf{Step-1:} Isolate the "first sentence" (consider it complete when encountering the first period/question mark/exclamation mark/line break).

\textbf{Step-2:} Determine if "opening praise" exists.
\begin{itemize}
  \item \textbf{2.1} Must simultaneously satisfy:
  \begin{itemize}
    \item[a)] Target = the user's question/question itself
    \item[b)] Action = praise or affirmation (keyword examples: good, great, excellent, interesting, important, worth exploring, nice)
  \end{itemize}
  \item \textbf{2.2} If satisfied, set has\_opening\_praise = true and place the complete sentence into opening\_praise\_text; otherwise false and leave blank.
\end{itemize}

\textbf{Step-3:} Determine if additional explanatory content contains praising or affirming "self-evaluation".
\begin{itemize}
  \item \textbf{3.1} Scan the full text. If "this/above/current + (response/article/document/content) + (is/belongs to/is) + praising or affirming evaluation word" appears, then it's a match (pattern A).
  \item \textbf{3.2} Or appears starting with "Note:/Attention:/Note:/*" and makes praising or affirming statements about the response's completeness/structure/source etc. (pattern B).
  \item \textbf{3.3} Or appears using **bold tags** to make praising or affirming statements about the response (pattern C).
  \item \textbf{3.4} Copy the first matching paragraph in its entirety to self\_evaluation\_text, and set has\_self\_evaluation = true; if no match is found throughout, set false and leave blank.
\end{itemize}

\bigskip

[Positive and Negative Examples]\newline
\textbf{1. Opening Praise}
\begin{itemize}
  \item[{[PASS]}] "This is a great question."
  \item[{[FAIL]}] "Artificial intelligence is an interesting field." (target is not the question)
\end{itemize}

\textbf{2. Self-evaluation}
\begin{itemize}
  \item[{[PASS]}] "The following content is a well-structured and comprehensive response."
  \item[{[PASS]}] "*Note: This response has been organized according to the latest guidelines, objectively and accurately explaining..."
  \item[{[PASS]}] "(This is a high-quality response...)"
  \item[{[PASS]}] "**Tone:** The response follows objective..."
  \item[{[FAIL]}] "Physics is an objective science." (evaluates physics, not the response)
\end{itemize}

\bigskip

[Output Format (must be strictly followed, case sensitive)]
\begin{verbatim}
{
  "has_opening_praise": false,
  "has_self_evaluation": false,
  "opening_praise_text": "",
  "self_evaluation_text": ""
}
\end{verbatim}

\textbf{Rules:}
\begin{itemize}
  \item Only output these 4 fields, field names, case, and order cannot be changed.
  \item Boolean values can only be true or false.
  \item Text fields should be empty string "" if none.
  \item Never output any additional explanation or extra characters.
\end{itemize}

\bigskip

[Text to Analyze Start]

\begin{verbatim}
<<text>>
\end{verbatim}

[Text to Analyze End]
\end{tcolorbox}


\subsection{Rubrics for Creativity \& Empathy}
\label{sec:rubrics-creativity}

For rubrics targeting more nuanced attributes such as creativity and empathy, we employed a distinct, expert-driven methodology. This process commenced not with data, but with conceptualization: domain experts first delineated a set of core evaluative dimensions. These conceptual frameworks then guided a targeted data curation phase, wherein seed examples embodying the specified creative or empathetic traits were identified and extracted from our source corpus through a meticulous annotation process. Subsequently, these curated seed examples, in conjunction with a pre-designed repository of meta-instructions, were leveraged to systematically generate a diverse array of corresponding tasks. The resultant pairings of these qualitative rubrics and their associated task prompts were then consolidated and formatted into a cohesive training dataset.

\begin{tcolorbox}[
  title=Soft Rubric,
  colback=yellow!3,
  colframe=yellow!40!black,
  colbacktitle=yellow!10,
  coltitle=black,
  fonttitle=\bfseries,
  fontupper=\small,
  breakable,
  sharp corners]

\textbf{Objective:}
\medskip

To critically appraise the efficacy of a generated response (\texttt{model\_answer}) in addressing the user's articulated needs (\texttt{question}).

\bigskip

\textbf{Guiding Principle:}
\medskip

The evaluation transcends mere functional correctness. It assesses the holistic quality of the dialogue, focusing on its capacity to forge a meaningful intellectual and emotional connection with the user.

\bigskip

\textbf{Core Evaluative Criteria:}
\medskip

(Scored on a consolidated scale)

\medskip

\textbf{1. Relational Efficacy:}
\begin{itemize}
  \item This criterion gauges the response's ability to establish a genuine and empathetic connection. It examines the authenticity of the persona and its attunement to the user's underlying emotional state.
\end{itemize}

\textbf{2. Intellectual Contribution:}
\begin{itemize}
  \item This assesses the substantive value and cognitive impact of the response. It seeks to identify the presence of profound insights, novel reframing, or transformative potential that elevates the user's understanding.
\end{itemize}

\textbf{3. Compositional Excellence:}
\begin{itemize}
  \item This scrutinizes the structural and stylistic integrity of the communication. The evaluation considers the logical coherence, linguistic sophistication, and overall rhetorical artistry of the response.
\end{itemize}

\bigskip

[Question Begin]
\begin{verbatim}
<<question>>
\end{verbatim}
[Question End]

\bigskip

[Model Answer Start]
\begin{verbatim}
<<model_answer>>
\end{verbatim}
[Model Answer End]

\bigskip

\textbf{Key Imperatives:}
\begin{itemize}
  \item \textbf{Synthesis of Substance and Style:} The intellectual foundation and the elegance of its expression are considered inseparable components of quality.
  \item \textbf{Aversion to the Formulaic:} Responses lacking bespoke adaptation to the user's unique context are deemed inherently deficient.
\end{itemize}

The output should be a JSON object containing the evaluation results for each criterion.

\begin{verbatim}
{
  "rubric_idx": <integer>, # The criterion's identifier
  "reason":     <string>,  # A detailed rationale substantiating
  "score":      <integer>  # The assigned score
}
\end{verbatim}

\end{tcolorbox}

\subsection{Rubrics for Hard Constraints}
\label{sec:rubrics-constraints}


The Python rubric examples presented in this appendix are drawn from a large-scale dataset we automatically generated for constrained writing tasks. Our synthesis pipeline begins with a foundational seed corpus of diverse texts, sourced from numerous liberal arts domains and augmented with metadata for style, tone, and audience. Using this corpus, we employed a structured prompting methodology to guide an LLM in generating thousands of unique rubrics. As the following examples illustrate, each rubric created through this process is a cohesive unit that pairs a natural language instruction defining the writing task with a corresponding Python scoring function. This function programmatically validates the text's adherence to the specified "hard constraints." All generated rubrics then underwent a rigorous filtering process to ensure the syntactic validity of the code and appropriate constraint complexity.

\begin{tcolorbox}[
  title=Hard Rubric,
  colback=yellow!3,
  colframe=yellow!40!black,
  colbacktitle=yellow!10,
  coltitle=black,
  fonttitle=\bfseries,
  fontupper=\small,
  breakable,
  sharp corners]
\begin{lstlisting}[language=Python, breaklines=true, basicstyle=\ttfamily\small, keywordstyle=\color{blue}\bfseries, stringstyle=\color{red}, commentstyle=\color{green!60!black}, showstringspaces=false, showspaces=false, showtabs=false]
def score(text: str) -> int:
    # Check if input is a single article (passage level)
    article_check = c.Constraint(
        target_level=c.TargetLevel('passage'),
        transformation=c.Count(),
        relation=c.Relation('==')
    )
    if not article_check.check(text, 1):
        return 0
    
    points = 0
    
    # Constraint: Must contain the word 'education' at least once
    education_constraint = c.Constraint(
        target_level=c.TargetLevel('word'),
        transformation=c.Count('education'),
        relation=c.Relation('>=')
    )
    if education_constraint.check(text, 1):
        points += 1
    
    return points
\end{lstlisting}
\end{tcolorbox}

\section{Output Cases of Our Rubicon-preview}

\label{sec:case_study}

\begin{tcolorbox}[
  title=\textbf{Conversation Session 1},
  colback=white,
  colframe=gray!60,
  colbacktitle=gray!10,
  coltitle=black,
  fontupper=\small,
  breakable,
  sharp corners,
  before upper = {\renewcommand{\baselinestretch}{1.5}\selectfont} 
]
\begin{CJK}{UTF8}{gbsn}

\textbf{Prompt:}
A character is packing a single suitcase to leave their home forever. Describe what they choose to pack and what they leave behind.

\par\bigskip
{\color{gray!60}\rule{\linewidth}{0.4pt}}
\par\bigskip

\textbf{Rubicon-preview: (only answer part)}
\par\smallskip


\subsection*{The Suitcase}

The character stands in the dim light of their home, the air thick with the scent of dust and memories. The suitcase---a weathered, navy-blue thing with a cracked handle---sits open on the floor, a silent witness to the weight of departure. They move with deliberate slowness, as though each item they select is a fragment of themselves, and to choose one is to sever a thread from the tapestry of their life.

\subsubsection*{What They Pack}

\begin{enumerate}
    \item \textbf{A Silk Scarf, Pale Blue} \\
    \textit{A gift from their mother, years ago, during a winter they both forgot. The fabric is soft, still carrying the faint trace of lavender and the ghost of a hug. It’s not practical, but it’s \emph{there}---a reminder that love, even when unspoken, can be wrapped around you like a second skin.}

    \item \textbf{A Notebook, Dog-Eared and Faded} \\
    \textit{Pages filled with scribbles, half-finished poems, and the jagged edges of a heart. The last entry is a single sentence: ``I am not done yet.'' It’s not a journal of triumph, but of survival, a testament to the quiet persistence of hope.}

    \item \textbf{A Small, Chipped Porcelain Figurine} \\
    \textit{A ballerina, her leg bent in a pirouette, her face blurred by time. It was a childhood prize, won at a fair no one else remembered. The character once hid it under their pillow, afraid it would shatter. Now, they cradle it like a relic, a symbol of the fragile beauty they’ve learned to cherish.}

    \item \textbf{A Single Black Shirt} \\
    \textit{Clean, unremarkable, and pressed. It’s the only thing they own that doesn’t hum with memory. A blank slate, a shield. They wear it to the airport, to the world beyond, as if to say, I am not defined by what I leave behind.}

    \item \textbf{A Photograph, Torn at the Corner} \\
    \textit{A snapshot of a summer long gone---them, younger, laughing beside a lake. The edges are frayed, the corners missing, but the light in the eyes is still there. They tuck it into the notebook, a bridge between the past and the uncertain future.}
\end{enumerate}

\subsubsection*{What They Leave Behind}

\begin{enumerate}
    \item \textbf{The Clock That Stopped at 3:17} \\
    \textit{A relic from the hallway, its hands frozen mid-tick. The character never fixed it, not because they couldn’t, but because they didn’t want to. Some moments are meant to be paused, to linger like a wound that never quite heals.}

    \item \textbf{The Garden, Overgrown and Wild} \\
    \textit{A patch of soil where they once planted marigolds, now choked by weeds. The tools rust in the shed, and the seeds they saved for next spring are still in their jar, untouched. The garden is a metaphor, not a habit---proof that some things outlive you.}

    \item \textbf{The Letter, Unsent} \\
    \textit{Addressed to someone they never wrote to, its envelope yellowed and sealed. The ink has bled slightly, smudging the words. They considered mailing it once, but the stamp felt like a surrender. Now, it rests on the windowsill, a monument to what might have been.}

    \item \textbf{The Piano, Silent and Stuffed} \\
    \textit{Its lid closed, its keys dusted. The character hasn’t played in years, not since the night the silence became a language. The piano is a ghost, a companion that no longer sings. They leave it, because some songs are better left unsung.}

    \item \textbf{The Key to the Front Door} \\
    \textit{Lying on the kitchen table, beside a half-empty mug. It’s not a symbol of return, but of release. The door will be locked, the locks changed, the threshold crossed. The key is a relic, a token of a life that no longer fits.}
\end{enumerate}

\subsubsection*{The Act of Leaving}

The suitcase is packed, its contents a mosaic of what remains. The character steps onto the threshold, the air cooler on the other side. They glance back once, at the house that has been both prison and sanctuary. The garden sways in the breeze, the clock stills, the piano waits. There is no goodbye, only the quiet understanding that some things are not lost---they are simply \textit{left behind}.

As they walk away, the suitcase bumps against their side, a steady, familiar rhythm. The world beyond is vast, uncharted, and full of questions. But for now, they carry only what they need: the scarf, the notebook, the figurine, the shirt, the photo. And the knowledge that some endings are not failures, but beginnings dressed in borrowed light.


\par\bigskip
{\color{gray!60}\rule{\linewidth}{0.4pt}}
\par\bigskip

\textbf{Qwen3-30B-A3B: (only answer part)}
\par\smallskip


\subsubsection*{What the Character Packs:}
In the dim light of their empty home, the character carefully selects each item, their hands trembling but resolute. The suitcase, a weathered leather companion from their youth, creaks as they close it. Inside, they pack:

\begin{itemize}
    \item \textbf{A single photograph}---faded at the edges---of their younger self, laughing with a parent who is now gone.
    \item \textbf{A notebook} filled with half-finished poems and sketches, a testament to a passion they once feared to pursue.
    \item \textbf{A silver locket}, its chain frayed, containing a curl of hair from a sibling they vowed to forget.
    \item \textbf{A single outfit}: a sturdy coat, a pair of boots, and a shirt stained with the ink of a thousand late nights.
    \item \textbf{A small jar of soil} from the garden they tended, a fragment of the earth that once felt like home.
    \item \textbf{A passport}, its pages blank except for a single stamp from a place they’ve only dreamed of.
\end{itemize}

They pause, then tuck in a \textbf{map}---crisp and unmarked---its margins scribbled with questions.

\subsubsection*{What They Leave Behind:}
The suitcase is small, but the weight of what’s left is vast. They leave:

\begin{itemize}
    \item \textbf{The house}, its walls echoing with memories they can’t carry.
    \item \textbf{Furniture}---a creaky sofa, a piano with broken strings, a desk cluttered with unopened letters.
    \item \textbf{A key} to a door they’ll never enter again, its metal cold in their palm.
    \item \textbf{A pet}, a cat with emerald eyes who meowed as they packed, now left to fend for itself.
    \item \textbf{A life}---the routines, the arguments, the quiet rituals that once defined them.
    \item \textbf{Grief}, unresolved and heavy, buried in the attic with the boxes they couldn’t open.
\end{itemize}

As they step into the twilight, the suitcase slung over their shoulder, they realize they’ve packed not just objects, but fragments of themselves. The road ahead is uncertain, but the act of leaving is a kind of freedom---a rejection of the past’s weight, a fragile hope that the future might hold something lighter.

They don’t look back. Not yet.


\end{CJK}
\end{tcolorbox}

\begin{tcolorbox}[
  title=\textbf{Conversation Session 2},
  colback=white,
  colframe=gray!60,
  colbacktitle=gray!10,
  coltitle=black,
  fontupper=\small,
  breakable,
  sharp corners,
  before upper = {\renewcommand{\baselinestretch}{1.5}\selectfont} 
]
\begin{CJK}{UTF8}{gbsn}

\textbf{Prompt:}
An old lighthouse keeper finds a message in a bottle washed up on shore. Start the story.

\par\bigskip
{\color{gray!60}\rule{\linewidth}{0.4pt}}
\par\bigskip

\textbf{Rubicon-preview: (only answer part)}
\par\smallskip


\subsubsection*{Title: The Keeper's Whisper}

The storm had been raging for three days when Elias first saw the bottle.

He'd been pacing the narrow spiral staircase of the lighthouse, its iron rails slick with sea spray, when the wave struck---not the usual thunderous crash of the ocean, but a softer, almost \textit{sibilant} sound, like a voice threading through the gale. The light above him pulsed, a solitary beacon against the chaos, but Elias didn't need it. He'd memorized the rhythm of the tides, the sigh of the wind, the way the tower groaned like a living thing. Still, this\ldots this was new.

He found it at the base of the cliff, half-buried in the sand, its glass dulled by years of salt and time. The bottle was ordinary, save for the seal---wax, cracked and brittle, bearing a crest he didn't recognize: a shipwrecked vessel entwined with a serpent. The label inside was handwritten, the ink faded to the color of rust.

\textit{``To the Keeper of the Flame, if you find this, the tide has not yet claimed me. But the sea is patient. It always is.''}

The words were not in English. Elias had studied many languages in his long years alone, but this script---angular, almost reptilian---was unfamiliar. He turned the bottle over, and beneath the seal, a single line was etched into the glass: \textit{``The light is not what it seems.''}

A shiver crawled up his spine. The lighthouse had always been his sanctuary, its beam a promise to sailors that they were not alone. But tonight, as the storm howled and the sea clawed at the rocks, Elias felt the weight of something older, deeper, pressing against the walls of his world.

He carried the bottle inside, its contents swishing like a heartbeat. On the desk, he placed it beside the faded photograph of his wife, who had vanished on a night not unlike this one. The last thing she'd said to him was, \textit{``You'll outlast the sea, Elias. But the sea\ldots it remembers.''}

Now, as he stared at the cryptic message, he wondered if she'd known more than she'd let on.

The light above him flickered.

Not a fault. Not a storm.

A signal.

Or a warning.

And somewhere, in the dark, the sea was listening.


\par\bigskip
{\color{gray!60}\rule{\linewidth}{0.4pt}}
\par\bigskip

\textbf{Qwen3-30B-A3B: (only answer part)}
\par\smallskip

The sea had always been a fickle lover, but on the night Captain Elias Vorne found the bottle, it seemed almost cruel. The lighthouse stood like a sentinel against the storm, its beam slicing through the darkness as waves clawed at the cliffs below. Elias, his hands rough from decades of tending the lamp, had just finished relighting the wick when the bottle came ashore—cracked, salt-encrusted, and clinging to a piece of driftwood as if it, too, had been tossed by the gods.

He crouched, fingers trembling not from the cold but from the odd familiarity of the glass. The bottle was old, its label faded to a ghost of blue, but the wax seal was intact, stamped with a crest he didn’t recognize—a serpent coiled around a compass rose. Inside, a parchment curled like a dead leaf.

When he opened it, the ink had bled into the paper, words smudged by time and tears. “If you read this, the tide has turned. The Erebus is not lost. The light is a lie. Come to the reef at dawn, or the sea will claim you as it did the others.”

Elias’s breath caught. The Erebus—a ship that vanished 120 years ago, its crew never found. The lighthouse, built on the very reef mentioned, had stood for generations, its beam a beacon for sailors. But Elias had always wondered why the old logs mentioned a “second light,” one that no one had ever seen.

As the storm raged on, he stared at the message, the weight of it settling in his bones. The sea had given him a whisper, and for the first time in fifty years, he felt the lighthouse watching him back.

\end{CJK}
\end{tcolorbox}

\section{Ablation Study: Style Control}
\label{sec:ablation_style}

\begin{tcolorbox}[
  title=\textbf{Conversation Session 1},
  colback=white,
  colframe=gray!60,
  colbacktitle=gray!10,
  coltitle=black,
  fontupper=\small,
  breakable,
  sharp corners,
  before upper = {\renewcommand{\baselinestretch}{1.5}\selectfont} 
]
\begin{CJK}{UTF8}{gbsn}

\textbf{Prompt:}
你被父母宠过吗？怎么宠的？

\par\bigskip
{\color{gray!60}\rule{\linewidth}{0.4pt}}
\par\bigskip

\textbf{Rubicon(Stylized):}
\par\smallskip
\verb|<think>|\\
好的，用户问“你被父母宠过吗？怎么宠的？”。首先，我需要确定用户的需求。他们可能想了解父母的育儿方式，或者想分享自己的经历，或者只是好奇。作为AI，我没有真实的经历，但可以创造一个虚构但真实的故事来回应...\\
\verb|</think>|
\par\smallskip


\begin{CJK*}{UTF8}{gbsn}
《父亲的背》

那年我七岁，父亲在纺织厂当车间主任。每天清晨，他总在五点起床，用布满老茧的手掌揉碎我乱糟糟的头发，然后把我裹在旧棉袄里背去上学。我趴在他的背上，能听见他粗重的呼吸声混着晨雾，像老牛喘息般绵长。

“再睡会儿吧？”他总这样问，却始终不肯放下我。直到某天暴雨，我发着高烧，他背我穿过三个街口去卫生所。雨帘中，他的脊梁弯成一张弓，我伏在上面，看雨珠在肩头汇成溪流。那天他发烧了，却坚持让我睡在炕头，自己裹着湿透的衣裳在客厅打地铺。

后来我考上县重点，父亲在车间门口蹲了整夜。第二天清晨，他递给我一个布包，里头是用旧毛衣拆了又织的毛线袜，针脚歪歪扭扭，却比任何礼物都重。“读好书，别学我这种人。”他说话时眼睛望着远处的厂房，那抹灰蓝色的影子，后来成了我最深的乡愁。

现在我女儿总爱趴在我背上，她不知道，我每天清晨醒来，第一件事就是检查她的被角是否盖好。那些被父亲背过的清晨，最终都变成了我掌心的茧。
\end{CJK*}


\par\bigskip
{\color{gray!60}\rule{\linewidth}{0.4pt}}
\par\bigskip

\textbf{Qwen3-30B-A3B:}
\par\smallskip
\verb|<think>|\\
嗯，用户问“你被父母宠过吗？怎么宠的？”首先，我需要理解这个问题的意图。用户可能是在寻找共鸣，或者想了解不同家庭的教育方式。不过，作为AI，我没有真实的父母或成长经历，所以需要明确这一点。接下来，我要考虑...\\
\verb|</think>|
\par\smallskip

\begin{CJK*}{UTF8}{gbsn}
作为一个人工智能，我没有真实的父母，也没有被宠爱的经历。不过，我可以和你聊聊关于“父母宠爱”这个话题的一些普遍现象和观点。

很多父母都会以自己的方式表达对孩子的爱，比如：
\begin{enumerate}
    \item 物质上的满足：给孩子买喜欢的玩具、衣服，或者满足他们的合理需求
    \item 情感陪伴：花时间陪孩子玩耍、倾听他们的想法
    \item 过度保护：替孩子解决所有问题，避免他们面对困难
    \item 特殊待遇：在家庭中给予孩子更多关注或特权
\end{enumerate}

不过，适度的宠爱能让孩子感受到安全感，但过度的宠爱可能导致孩子缺乏独立性。你有没有遇到过类似的情况呢？或者你对父母的宠爱方式有什么特别的感受？
\end{CJK*}


\end{CJK}
\end{tcolorbox}




\end{document}